\documentclass[letterpaper, 10 pt, conference]{ieeeconf}  % Comment this line out if you need a4paper

\IEEEoverridecommandlockouts                              % This command is only needed if 
\usepackage{graphics} % for pdf, bitmapped graphics files
\usepackage{epsfig} % for postscript graphics files
\usepackage{mathptmx} % assumes new font selection scheme installed
\usepackage{times} % assumes new font selection scheme installed
\usepackage{amsmath} % assumes amsmath package installed
\usepackage{amssymb}  % assumes amsmath package installed
\usepackage[T1]{fontenc}

\title{\LARGE \bf
A 3D-Deep-Learning-based Augmented Reality Calibration Method for Robotic Environments using Depth Sensor Data
}

\author{Linh K{\"a}stner$^{1}$\thanks{$^{1}$Linh K{\"a}stner and Jens Lambrecht are with the Chair Industry Grade Networks and Clouds, Faculty of Electrical Engineering, and Computer Science,
		Technical University of Berlin, Berlin, Germany}, Vlad Catalin Frasineanu$^{2}$\thanks{$^{2}$Vlad Catalin Frasineanu is with Microsoft Cloud and AI Security, Vancouver, Canada
 {\tt\small linhdoan@tu-berlin.de}} and Jens Lambrecht$^{1}$
}

\begin{document}

\maketitle
\thispagestyle{empty}
\pagestyle{empty}

%%%%%%%%%%%%%%%%%%%%%%%%%%%%%%%%%%%%%%%%%%%%%%%%%%%%%%%%%%%%%%%%%%%%%%%%%%%%%%%%
\begin{abstract}

Augmented Reality and mobile robots are gaining much attention within industries due to the high potential to make processes cost and time efficient. To facilitate augmented reality, a calibration between the Augmented Reality device and the environment is necessary. This is a challenge when dealing with mobile robots due to the mobility of all entities making the environment dynamic. On this account, we propose a novel approach to calibrate the Augmented Reality device using 3D depth sensor data. We use the depth camera of a cutting edge Augmented Reality Device - the \textit{Microsoft Hololens} for deep learning based calibration.
Therefore, we modified a neural network based on the recently published \textit{VoteNet} architecture which works directly on the point cloud input observed by the \textit{Hololens}. We achieve satisfying results and eliminate external tools like markers, thus enabling a more intuitive and flexible work flow for Augmented Reality integration. The results are adaptable to work with all depth cameras and are promising for further research.
Furthermore, we introduce an open source 3D point cloud labeling tool, which is to our knowledge the first open source tool for labeling raw point cloud data.

\end{abstract}

%%%%%%%%%%%%%%%%%%%%%%%%%%%%%%%%%%%%%%%%%%%%%%%%%%%%%%%%%%%%%%%%%%%%%%%%%%%%%%%%
\section{INTRODUCTION}

The need of a 3D understanding of the environment is an essential for tasks such as autonomous driving, Augmented Reality (AR) and mobile robotics. As the progress with 3D sensors intensified, sensors like light detection and ranging sensors (Lidar) can provide high accuracy for such use cases. 
Mobile robots are one of the main systems to profit from the recent progress in 3D computer vision research. In addition, their popularity increased due to their flexibility and the variety of use cases they can operate in. Tasks such as procurement of components, transportation, commissioning or the work in hazardous environments will be executed by such robots \cite{smartfactory} \cite{mobilerob}. However, operation and understanding of mobile robots is still a privilege to experts \cite{mobilerob2} as a result of their complexity. 
On this account, Augmented Reality (AR) has gained popularity due to the high potential and ability to enhance efficiency in human robot collaboration and interaction which had been proved in various scientific publications \cite{hashimoto}, \cite{heydari}, \cite{fang}.  
AR has the potential to aid the user with help of spatial information and the combination with intuitive interaction technology, e.g. gestures \cite{arsystems}. Our previous work focused on AR-based enhancements in user understanding for robotic environments. In  \cite{lambrecht} we simplify robot programming with the help of visualizing spatial information and intuitive gesture commands. In  \cite{kaestner} we developed an AR-based application to control the robot with gestures and visualize its navigation data, like robot sensors, path planing information and environment maps. Other work used AR for enhanced visualization of robot data or for multi modal teleoperation \cite{ar1} \cite{armultimodal}. 
The key aspect to facilitate AR is the initial calibration between AR device and environment. State of the Art approaches are relying on marker detection which proves to be accurate but unhandy. Additionally, markers cannot be deployed everywhere especially not in complex dynamic environments for use cases like autonomous driving. For such use cases neural networks are increasingly considered.
Deep Learning algorithms have already had a great impact in 2D computer vision. While
the majority of projects are working on 2D image processing and object recognition, with
the recent advances in the depth sensor infrastructure, the 3D research has been intensified. Whereas most of the research are focused on \textit{Lidar} and their use cases for autonomous driving, within this paper, we will try to adopt these ideas and integrate them within an AR device
and its sensors.
More specifically, we rely on depth sensors of the AR device to train a neural network working solely on raw point cloud data making the operation more intuitive and flexible without any help from external tools like markers.
Our results can also be adapted for every other depth sensor system which opens many possibilities for further research. 
The main contributions of this work are following:
\begin{itemize}
	\item Proposal of a modified neural network for object detection based state of the art neural networks working directly on point cloud data
	\item Combination of neural network object detection with AR for improved initial calibration and facilitating more intuitive and flexible usage of AR
	\item Extension of an AR based robot control application from our previous work \cite{kaestner} for more flexible usage
	\item Development of an open source 3D annotation tool for 3D point clouds
\end{itemize}
The rest of the paper is structured as follows. Sec. II will give an overview of related work. Sec. III will present the conceptional design of our approach while sec. IV will describe the implementation. Sec. V will describe the training process and Sec VI will demonstrate the results with a discussion in Sec. VII. Finally, Sec. VIII will give a conclusion.

\section{RELATED WORK}

Due to the high potential of AR, it has been considered for integration into various industrial use cases.
The essential step to facilitate AR is the calibration between the AR device and the environment. 
Manual calibration using external tools or special equipment is the most straightforward method and requires the user to manually align 2D objects to objects in the real environment which position is known to the calibration system. 
 Azuma and Bishop \cite{azuma1994improving} propose a AR camera pose estimation by manually aligning fiducial like virtual squares to specific locations in the environment.
 Similar work from Janin et al. \cite{janin1993calibration} and Oishi et al. \cite{oishi1996methods} require the user to align calibration patterns to different physical objects which positions are known beforehand. However, manual calibration bear disadvantages in a continuously automated environment due to the necessity to have a human specialist to do the calibration. The external tools themselves can be expensive or are case specific.\\ Recent research focus on automated calibration methods rather than manual ones. Most ubiquitous are marker based approaches. Open source libraries and toolkits like \textit{Aruco} or \textit{Vuforia} make it possible to deploy self created fiducial markers for pose estimation. The marker is tracked using computer vision and image processing to detect the marker and generate the pose estimation for the camera. Especially when working with static robots, marker based approaches are widely used due to the simple setup and competitive accuracy. Several work including Aoki et al. \cite{aoki2018self}, Ragni et al. \cite{ragni2018artool} or \cite{chu2018helping} et al. relied on a marker based calibration between robot and AR device. 
However, Baratoff et al. \cite{baratoff2002interactive} stated, that the instrumentation of the real environment, by placing additional tools like markers, sensors or cameras is a main bottleneck for complex use cases especially in navigation that are working with dynamically changing environments. Furthermore, deploying additional instruments require more time to setup and are not able to react to changes within the environment. Especially for industrial scenarios this is the main concern \cite{baratoff2002interactive}. \\
On this account, marker-less approaches have become a major focus \cite{yan2017application}. This includes the so called model based approaches which use foreknown 3D models to localize objects and generate the pose estimation for the AR device. This is done using computer vision and frame processing to automatically match the known 3D models on the object from the incoming frame. Subsequently the pose is estimated. Works by Bleser et al. \cite{bleser2006online} were able to establish a competitive approach using CAD models and achieving high accuracy. 
 The main bottleneck is, that these approaches are very resource intensive due to the processing of each frame.
 In addition, these methods are highly dependent on the CAD models, which are not always available. Furthermore, the approach becomes less flexible, which is crucial in dynamic environments for tasks such as autonomous navigation of mobile robots. \\
 With the recent advances in machine learning algorithms, especially deep learning, and the higher computational capabilities, networks can be trained to recognize 3D objects.
 An overview of current research is provided by Yan et al. \cite{yan2017application}.  Garon and Lalonde \cite{garon2017deep} proposed an optimized object tracking approach with deep learning to track 3D objects with Augmented Reality. 
 Alhaija et al. \cite{alhaija2018augmented} worked with neural networks for object detection and segmentation for AR-supported navigation. The researchers relied on 2D object detection approaches and one of the main drawbacks are less accuracy compared to 3D based approaches because the 3D pose has to be estimated solely with 2D information which leads to inaccuracy. 
Rapid development of 3D sensor technology has motivated researchers to develop efficient
3D based approaches and several networks have been proposed to work with 3D sensor information. 
There exists fusion approaches which require both, 2D and 3D sensor data to improve accuracy even further. The 2 sensor streams have to be synchronized and calibrated which is a not trivial tasks to do and require additional steps. 
For this reason, several works were proposed which try to work only with the 3D information. 
Contrary to 2D data, the 3D input data is not ordered and the huge amount of points which can be up to millions is computationally much more demanding. Some works dealing with a preprocessing of the unordered 3D point cloud data to an ordered data structure. These are either based on \textit{Voxels} as done by the work of Zhou et. al proposes with \textit{VoxelNet} \cite{zhou2018voxelnet} or on a conversion of the point cloud into a birds-eye-view as Simon et. al proposes in ComplexYolo \cite{simon2018complex}.
However, quantization and preprocessing of the point cloud increases the risk of loosing important features and information. 
Another popular solution is based on the \textit{PointNet} network (or its updated version
PointNet++) proposed by Qi et. al \cite{qi2017pointnet} which is able to learn semantic segmentation and 3D object localization directly
from raw point clouds. Since the features are extracted directly from the raw representation,
without any preprocessing, there is no risk in loosing information, thus making this
method much faster and accurate.
As follow-up work, several networks were recently built on top of this network which
are using it as the backbone for feature learning. Two of them are \textit{VoteNet} from Qi et. al \cite{qi2019deep} or \textit{PointRCNN} proposed by Shi et. al \cite{shi2019pointrcnn},
which showed promising results on benchmarking on popular data sets. They improve the
extracted features by considering the local structures in point clouds.

\section{CONCEPTION DESIGN}

\subsection{AR Device Calibration}
\begin{figure}[!h]
	\centering
	\includegraphics[width=3.3in]{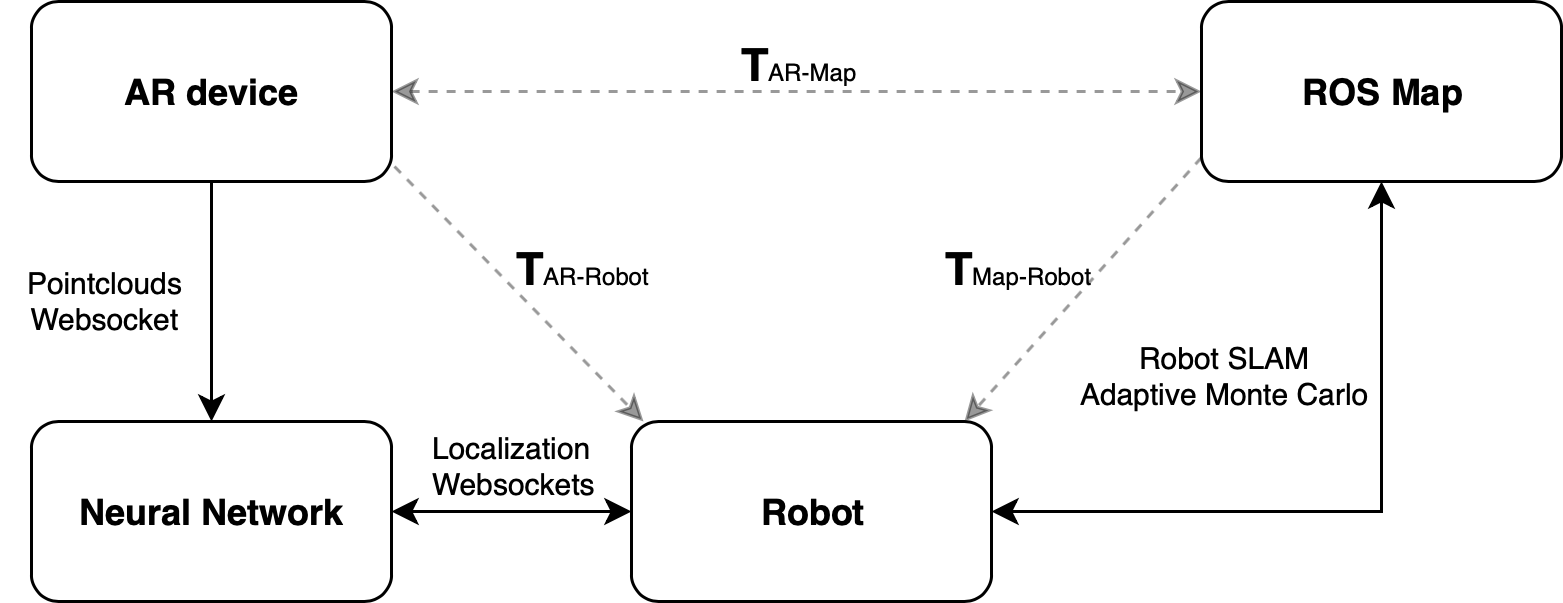}
	\caption{\textbf{Conception of Coordinate System Alignment }- To realize AR, the transformation between the AR device and the ROS map is to be considered ($T_{AR-Map}$), because all data e.g. sensor information, map and navigation data are with reference to the ROS map. The position of the robot within the ROS map is known via the internal robot SLAM ($T_{Robot-Map}$). Hence, to achieve a transformation between AR device and ROS map, we have to achieve a transformation between AR device and robot. This is done with our proposed deep learning method.   }
	\label{conceptcoord}
\end{figure}
The objective of this paper is to deploy a marker-less calibration method for AR device and robot using depth sensor data.
For our case, due to the mobility of our robot, a continuous localization of the robot is to be ensured. In addition, the AR device is fully mobile as well, which makes both entities able to move freely and a single marker tracking approach like used in similar works insufficient. For the AR application first developed in our previous work, we proposed a two-stage calibration between AR headset and robot. First, the \textit{Hololens} is calibrated with the robot using \textit{Aruco} markers. Second, the spatial anchor capability of the \textit{Hololens} is used for calibration of the robot with the Robot Operating System (ROS) map. 
The robot map is the general reference map from which all sensor data and information are displayed. Thus, calibration between the AR device and the ROS internal map is the overall goal to visualize data properly within the AR headset. The robot position is always known by the map through the robot SLAM packages (e.g. \textit{Adaptive Monte Carlo}). The position of the \textit{Hololens} as an external tool however, must be integrated into ROS. 
For detailed technical information we refer to our previous work \cite{kaestner}.  This work will replace the initial calibration (first stage) of the AR device and the robot, done previously with a marker based approach. This is done by using a neural network working directly on the point cloud sensor input of the \textit{Hololens}. The conception design is illustrated in Fig. \ref{conceptcoord}.

\subsection{Neural Network Selection}

We are driven by the release of state of the art neural networks working directly on raw point cloud data rather than preprocessed point clouds. Furthermore, improving depth sensor technologies and the recent release of the \textit{Microsoft Research Mode} provide easy to access raw sensor data to work with. On this account, we use the \textit{Hololens} internal depth sensor to extract point clouds for our neural network to detect the robot.
As stated in the related work section, suitable approaches are \textit{VoxelNet}, \textit{VoteNet} and \textit{PointRCNN} which operates directly on raw pointcloud data without the need of potentially harmful preprocessing steps of the sensor information. 
\textit{VoxelNet} and \textit{PointRCNN} both contain a large amount of layers making the network too complex for our use case. Additionally, for demonstration purposes we will operate with a relatively small dataset consisting of less then 1000 point clouds. Given the complexity of those networks, overfitting is a main concern. \textit{VoteNet} on the other site is a simple architecture based on the \textit{Hough Voting} strategy in point clouds. It is end-to-end trainable and the contains a comfortable amount of hyper parameters which makes the training process easier compared to first mentioned architectures. Furthermore, \textit{VoteNet} achieved remarkable results on challenging datasets. Therefore, our implementation is based strongly on the \textit{VoteNet} architecture. For our specific use case, modifications have to be done. These are explained in detail in the implementation chapter IV.

\subsection{Hardware Setup}
The hardware setup contains a mobile robot and a head mounted AR device - the \textit{Microsoft Hololens}. We are working with a \textit{Kuka Mobile Youbot} running ROS \textit{Hydro} on a \textit{Linux} version 12.04 distribution. As a development environment for the \textit{Hololens}, we use a windows notebook with \textit{Unity3D 2018} installed on which the application was developed. The sensor data acquisition is done with C++ code within \textit{Visual Studio 2019} and \textit{Microsoft Research Mode} for \textit{Hololens} (RS4 Update, April.2018). 
Both entities, robot and \textit{Hololens}, are connected to the same network and communicate via a \textit{WLAN} connection. 

\subsection{Overall Workflow}
\begin{figure}[!h]
	\centering
	\includegraphics[width=3.3in]{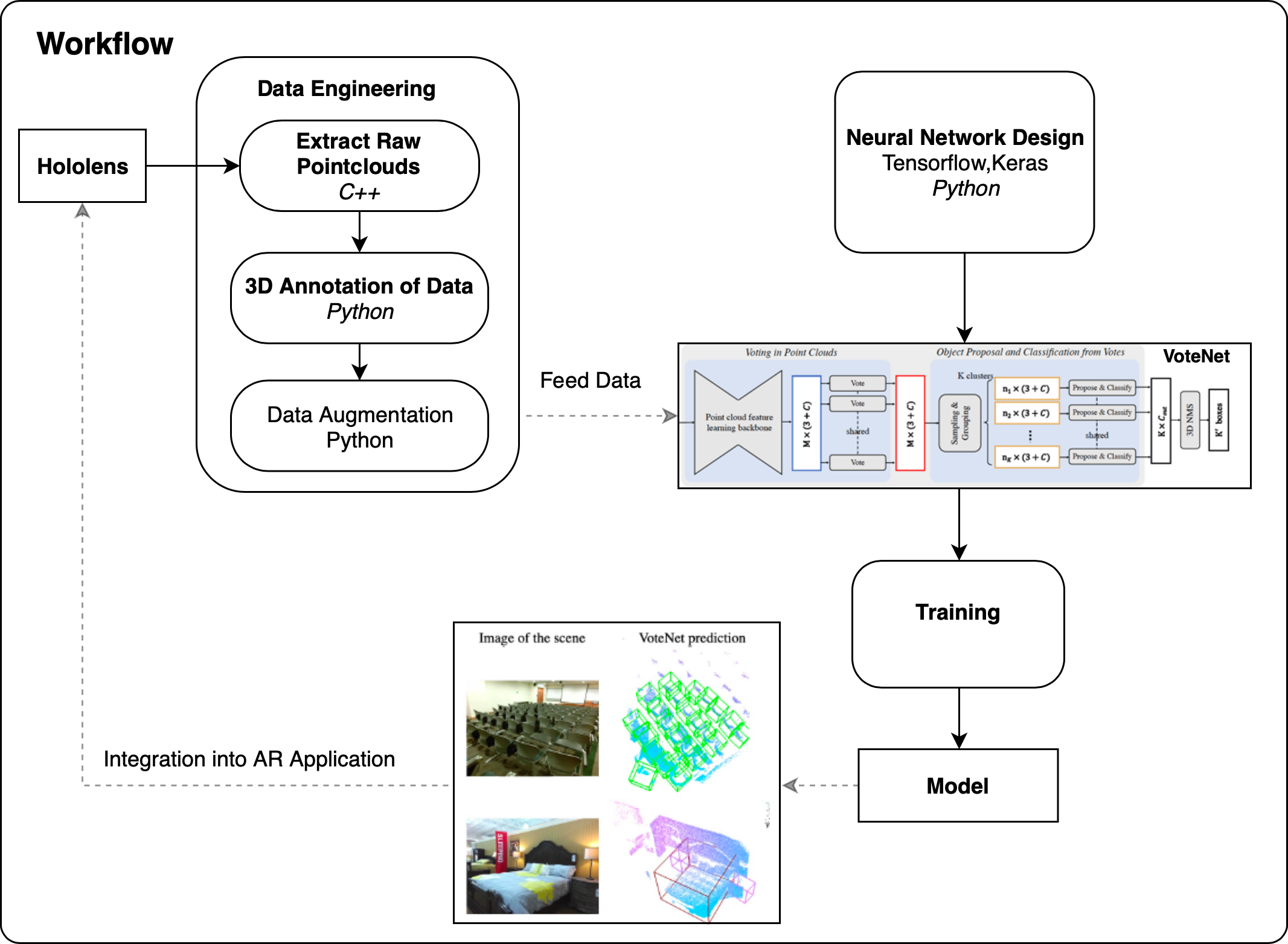}
	\caption{Overall Workflow}
	\label{ow}
\end{figure}

The overall workflow of our project is depicted in Fig. \ref{ow}.

After data extraction with the Hololens Research Mode, the point cloud data has to be annotated. A data augmentation step is included to enlarge our dataset and prevent overfitting. After training of the neural network, integration into the existing AR robot control application is to be achieved. Since all entities are connected through the same network, integration can be established via websocket connections that facilitates data exchange between the entities. The neural network model is deployed at the local \textit{GPU} server.

\section{IMPLEMENTATION}
This chapter will present the implementation part of each module in detail. 
\subsection{Pointcloud Data Extraction and Processing}
The point clouds are acquired with the \textit{Microsoft Research Mode} and the related open source library \textit{HololensForCV}, which contains a depth streamer project. We implement a new streamer on top of that, that converts the depth data into 3D point clouds which are saved for each frame inside the \textit{Hololens}. 
The point cloud processing takes 2-3 seconds which is why we are able to store a point cloud representation every 4 frames.
The conversion from depth data to a point cloud was developed using the camera intrinsics
parameters provided by the \textit{HololensForCV} library for the depth sensor. The depth streamer returns a depth buffer array $D$, which contains the distances for each pixel from the 2D image plane. We iterate through each pixel point of the 2D image, map it to the camera unit plane and normalize the results based on following equations.
To calculate a 3D point out of the 2D coordinated [u,v] following equations are used. 
\begin{equation}
[X,Y,Z] = Z \cdot [u, v, 1] =  [Z\cdot u, Z \cdot v, Z]
\end{equation}
where Z is the normalization of the distance array D which is computed using equation (2). 
\begin{eqnarray}
Z = \frac{D}{\sqrt{(u^2+v^2+1)}} \\
X = \frac{D \cdot u}{\sqrt{(u^2+v^2+1)}} \\
Y = \frac{D \cdot v}{\sqrt{(u^2+v^2+1)}} 
\end{eqnarray}
This is done for each pixel from the 2D image.
We worked with the long throw depth sensor of the \textit{Hololens} providing points with a maximum distance of 4 meters. Thus, we considered only the depth values between 0.4  and 4 meters as valid values. For each frame that we process, we store its corresponding 3D point cloud representation.

\subsection{3D Data Annotation}

The next step in the pipeline is the annotation of the acquired point cloud data. To the time of this work, no reliably, open source tool could be found dealing with the direct annotation of 3D point cloud. On this account, we propose a 3D point cloud labeling tool for annotating point cloud data. Detailed instructions and explanations on how to use it can be found on Github. In the following, the tool will be explained briefly.
The tool takes as input raw point cloud data which will be visualized with the python library \textit{PPKT}. In order to annotate the robot in the visualizer, the user has to select 3 points that represent
3 corners of the base of the robot, which are used by the tool to compute the
fourth corner together with the base plane on which the robot is situated. Each point
inside the point cloud is projected onto the base plane. 
All points situated inside the calculated base are considered as part of the object.  Out of these points, the highest one which is smaller than a predefined threshold is picked and the coordinates will be used to compute the height of the bounding box. The center is computed by choosing the point
on the normal of the base plane that starts from its center, it is at a distance of height/2 from
the base plane and it is situated inside the robot.
The tool goes through each point cloud, renders it, waits for the user to select the robot inside the
visualizer by choosing 3 points, and then computes the center of the bounding box corresponding to that selection
together with its size and rotation according to above explanations. 
\begin{figure}[h!]
	\centering
	\includegraphics[width=3.4in]{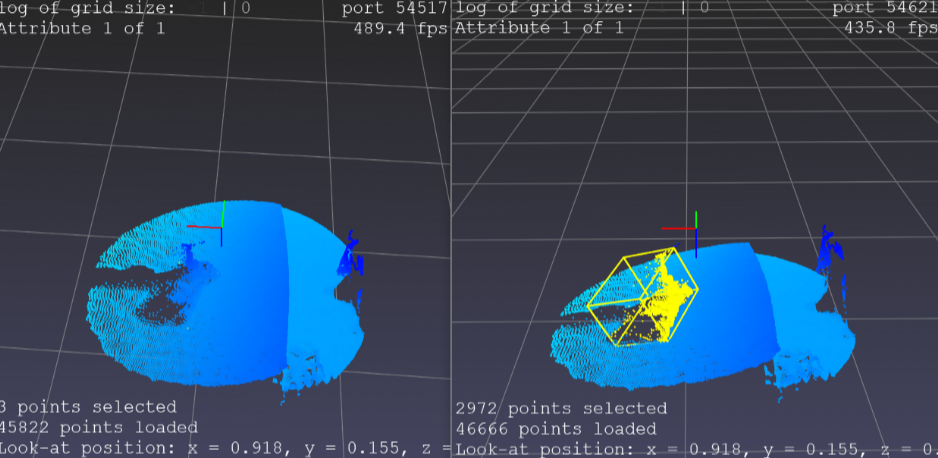}
	\caption{\textbf{Labelling tool}
		- The labelling tool built using the \textit{PPTK} visualizer. On the left side
		 the initial visualization where the user has to select 3 corners of the base of the robot is shown.
		On the right side the processed bounding box that the user has to confirm is displayed. If something is not correct, then the user can retry the selection of the corners.}
	\label{tops}
\end{figure}

%\begin{figure}[h!]
%	\centering
%	\includegraphics[width=3.4in]{bbox}
%	\caption{\textbf{The resulting bounding box} - In total we have labeled over 800 pointclouds manually. If applying data augmentation on the dataset, the bounding box coordinates have to be augmented as well. Detailed instructions and code to achieve this can be found in the github project}
%	\label{tops}
%\end{figure}

\subsection{Neural Network Design}
\begin{figure}[h!]
	\centering
	\includegraphics[width=3.4in]{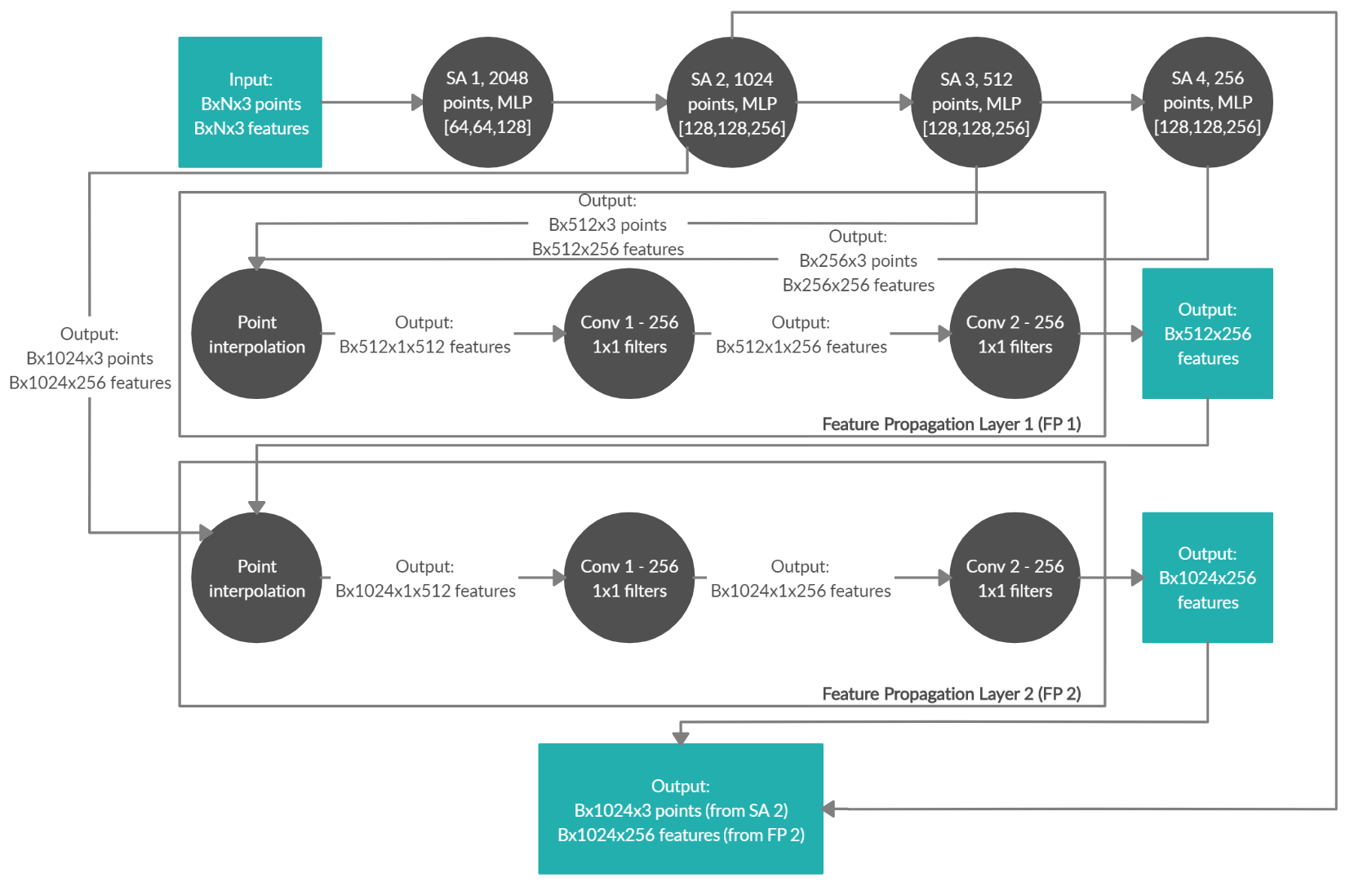}
	\caption{\textbf{Feature Learning Module} - The input goes through 4 Set Abstraction layers (SA) and are concatenated together by two feature propagation layers (FP). In the end, the points sampled by the second SA represent the final seed points while the output of the last FP layer represent their corresponding features. The output is then further processed in the voting module which is presented in the next section }
	\label{fl}
\end{figure}

\begin{figure}[h!]
	\centering
	\includegraphics[width=3.4in]{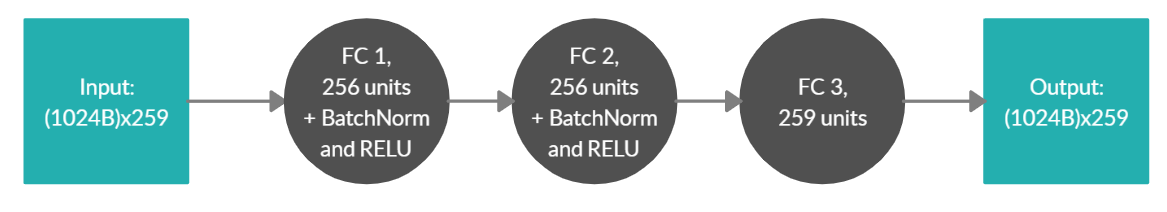}
	\caption{\textbf{Voting Module} - The input that comes from the previous
		module is concatenated together and reshaped into a 2D
		matrix with sizes (1024B)x259 where B represents the number of point clouds in the batch. Then
		the input is further processed by 3 fully connected layers (FC) together with a
		batch normalization and RELU activation step for the first two layers. The shape of the output will be
		the same as the input and it will represent the offset that has to be added to the seeds in order to
		get the votes. In the end Bx1024x3 votes (XYZ positions) and Bx1024x256 features will be outputted}
	\label{vm}
\end{figure}

\begin{figure}[h!]
	\centering
	\includegraphics[width=3.4in]{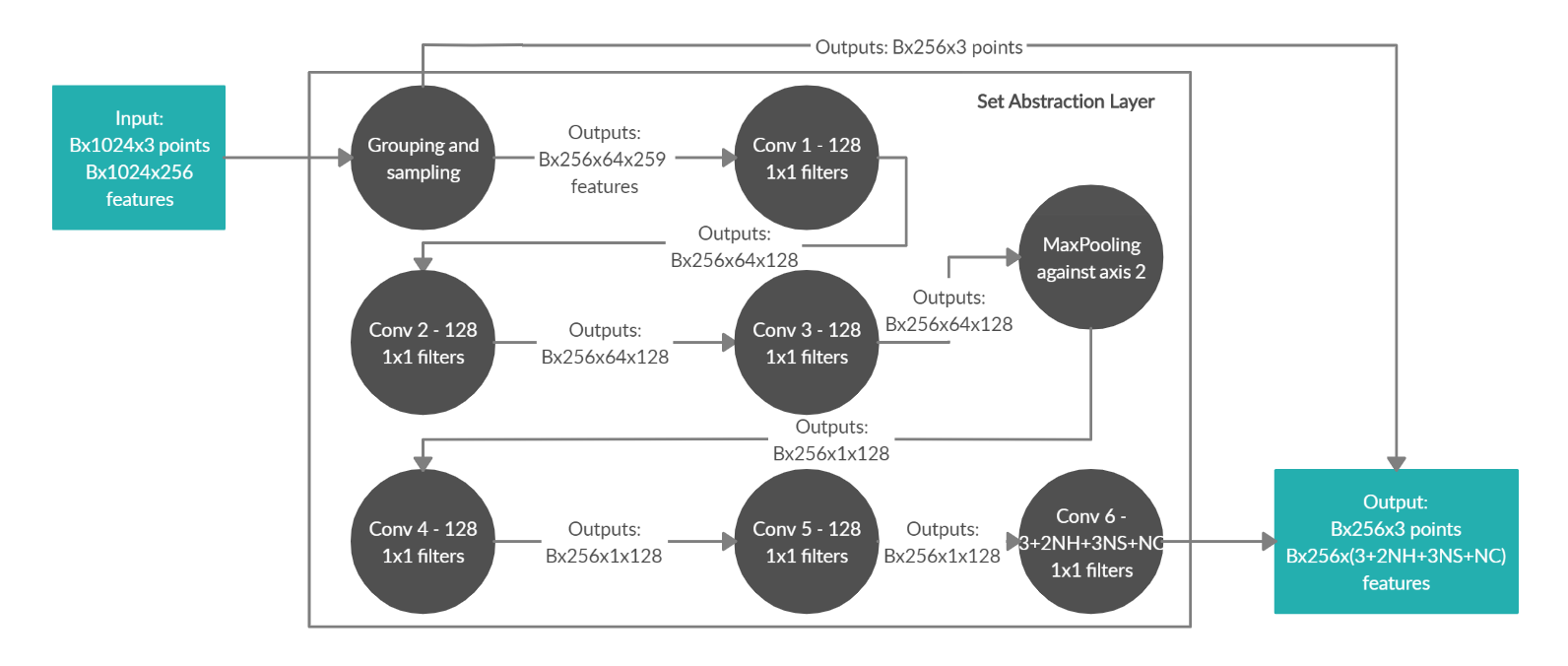}
	\caption{\textbf{Object Proposal Module} - The Set Abstraction layers (SA) will sample only 256 points from the votes that will form groups of length with 64 point. This is processed along with another final multilayer perceptron (MLP) with 3 layers to generate the proposals) Each point in the final output will contain }
	\label{opm}
\end{figure}

In this section, the exact architecture together with the modifications that were made to the original network are presented in detail. The purpose of the
original network is to find all objects existing in a scene and then label them between different
existing classes. Therefore, the original architecture learns to vote to the closest existing object. Since our use case require the detection of an specific object, the mobile robot, we modify the architecture to learn voting only towards the center of
the robot if it exists in the current input. This was achieved by making changes to the original
loss function that we are going to discuss in this chapter.
The network can be divided into 3 main modules which are adapted from the original paper. We have  visualized the architecture of each for better understanding in Fig. \ref{fl}, Fig. \ref{vm} and Fig \ref{opm}.\\
\textbf{1) Feature learning module.}
As a first step, the network has to learn local geometric features
and sample the seed points. For that, 4 set abstraction layers are applied followed by two feature propagation layers. These layers act as the backbone network and are adapted from the \textit{Pointnet++} architecture. After each fully connected layer, a batch normalization step was applied in order to keep the outputs in similar scales.\\

\textbf{2) Voting Module} After the feature learning stage, 1024 seed points will be distributed over the input pointclouds. For our case the point cloud contains between 30k-40k points. The seed points will learn the behavior of voting for the center of an object. This is closely supervised by a loss function $L_{vote-reg}$ which is defined as 
 \begin{equation}
L_{vote-reg} = \frac{1}{M_{pos}} \sum_{seed\,i\,is\,positive}^{max} |\Delta x_i - \Delta x^*_i|
\end{equation}
Where $\Delta x_i$ represents the offset values of seed points $x_i$. $\Delta x^+_i$ represents the ground truth offset that has to be applied to the position of each seed point in order to get to the position of the center of the ground truth bounding box. Then the positions of the votes will be $\Delta x_i + \Delta x_i *$. $\Delta x_i *$ represents the ground truth offset. The advantage of using this strategy is that votes are now not situated on the
surface of an object anymore (like the seed points) and all votes that are generated by seeds close to the object are will be close to each other which makes it easier to cluster
them and concatenate features from different parts of the object together.

\textbf{3) The Object Proposal Module}
The last part of the network will cluster the votes and concatenate their features together through a \textit{PointNet} set abstraction layer combined with another final MLP with 3 layers that generates the final proposals. The final output will contain 3+2NH+3NS+NC features where NH represents the number of heading classes, NS represents the anchors which is the size of the bounding box for a specific object. Since we have one class to detect, 3 NS values (length, width, height) are considered. NC is the number of classes to predict, which in our case is 1.

\subsection{Loss functions}
In this section the loss functions and their modifications are explained. Our overall loss function is 
\begin{equation}
L = 100L_{vote-reg} + L_{box} + 20L_{sem-cls}
\end{equation}

\subsubsection*{ \textbf{Center regression loss ($L_{vote-reg})$} }
Here offset of the votes of the
	seed points that are close/inside the existing object are minimized to make sure that the
	votes of these seed points will be situated closer to the ground truth center position of
	the bounding box.

\subsubsection*{\textbf{Semantic loss ($L_{sem-cls}$)}}

This loss was a modified using a version that was
	presented in \textit{VoxelNet} \cite{zhou2018voxelnet}. The loss is defined as following:
	\begin{equation}
	L_{sem-cls} = \alpha \frac{1}{N_pos} \sum_{i} L_ds (p_i^{pos},1) + \beta \frac{1}{N_{neg}} \sum_{j} L_{ds}(p_j^{neg},0)
	\end{equation}
	
Using this loss, we make sure that a seed point that is inside the object and required to
	vote for the center of the object has a confidence score close 1, while a seed point that
	is far away from the object and does not have to vote for the object has a confidence score close to 0.

\subsubsection*{\textbf{Box Loss ($L_{box}$)}}
 This loss function was modified by two small changes: we removed the size-cls loss since we have only one
	type of object and only on initial anchor. We directly predict the regression
	parameters for length, width and height that we need to apply in order to get from the
	initial anchor to the true size of the robot and the angle class loss since we use only
	one class for estimating the rotation angle and make the network directly predict the
	residual angle. The loss is defined as following:
	\begin{equation}
	L_{box} = 10 L_{center-loss} + 5 L_{heading - residualLoss} + 10L_{size-residualLoss}
	\end{equation}

\section{Training process}
The network was trained for 480 epochs (where a epoch means one pass through the whole
data set), using batches of size 8 and the Adam optimizer for minimizing the loss function. We
started with a learning rate of 0.001 for the first 200 epochs, decreased it by 10 for the next
200 and then in the end decreased it again by another 10 for the last 80 epochs. This process
takes approximately 12 hours on a \textit{Tesla K80} GPU. 
We split the initial data set which had 597 point clouds into 537 point clouds that we used for
training and 60 point clouds that were used for testing/evaluation of the network.
For training the size of each mini-batch must contain the same number of points. For that, 25000 points were randomly sub sampled from the input point clouds. For testing, we
used batches of size 1, so the initial number of points in each point cloud was kept. Other than
this, no other pre-processing step was necessary. During the training process, some basic data augmentation methods were used in order to
avoid rapid overfitting of the data set. These methods included:
\begin{itemize}
	\item Scaling of all points in the point cloud  by a random factor.
	\item Flipping the points against the X and Y axis. The corresponding coordinate of the center and to the rotation angle $\theta$ becomes 180-$\theta$
	\item Rotating all points around the z axis by a random angle
\end{itemize}

\section{Results} 

\subsection{Training Results}
\begin{figure}[h!]
	\centering
	\includegraphics[width=3.4in]{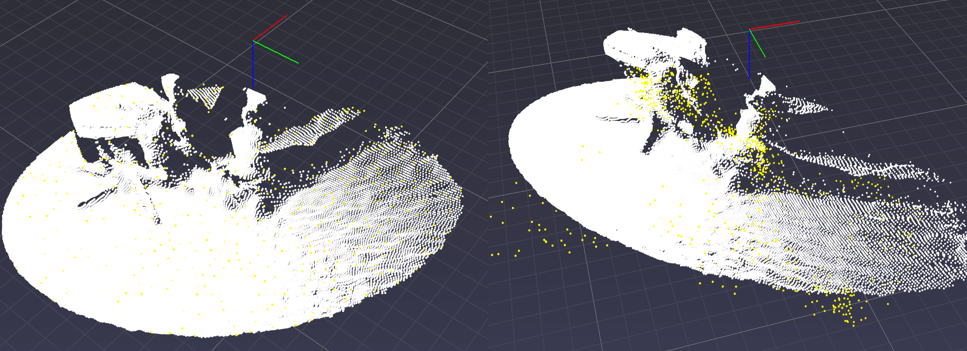}
	\caption{\textbf{Seeds and Votes on the training data} - the yellow dots represent the seeds and the votes generated throughout the learning process. It can be observed that the seeds will be randomly distributed over the point cloud. The vote points near the object cluster around the center of the object while seed points far away from the object learned to vote outside the point cloud to not influence the loss function and be eliminated later}
	\label{rp}
\end{figure}

In the following, the training results will be demonstrated.  Fig. \ref{rp} illustrates the training results on a point cloud input.
The two main steps of seeding and voting explained earlier can be observed clearly. The seed points, which are randomly distributed in the input point cloud, generate votes which learned to be either at the center of the object or outside the point cloud. Fig. \ref{rp} demonstrates that this behavior could be learned throughout the training process with the modified loss functions mentioned in chapter IV. Seed points near the object will generate votes which are placed near the center of the object while seed points which are far away from the object will vote for points inside the object point cloud. In the next step these will be eliminated with the clusters which are generated with the high votes as region proposals for object detection. 

\begin{figure}[h!]
	\centering
	\includegraphics[width=3.4in]{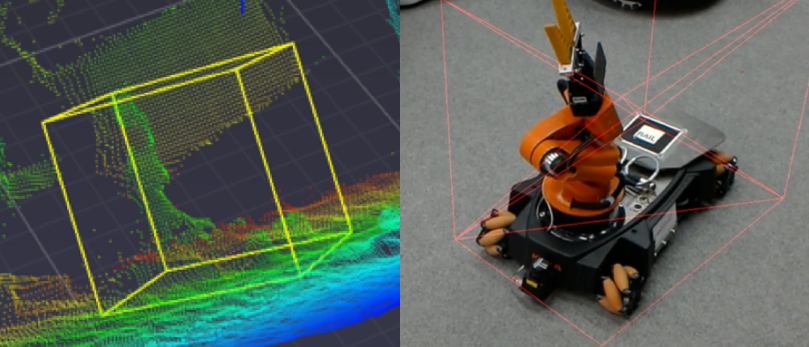}
	\caption{\textbf{Demonstration of the robot localization}. Left, from the visualizer. Right, from the \textit{Hololens} view: the point clouds are extracted and send to the neural network which makes the predication about position, size and rotation of the robot. Afterwards a 3D bounding box (red) is drawn directly at the user sight.}
	\label{demo}
\end{figure}

Fig. \ref{demo} demonstrates the localization via the \textit{Hololens} view. The incoming point cloud data, which will be recorded by looking at the robot is evaluated by the neural network which provides accurate results. The processing of the incoming point cloud data takes 4 seconds. Communication to the neural network and evaluation is depending on the Internet connection usage and takes around 1-2 seconds.

\section{Discussion}
The detection of the robot is working reliable but still there are still some limitations. Due to the long preprocessing of the incoming point cloud data, the user has to wait for at least 2 seconds without moving. else wise the neural network will evaluate on a different point cloud. Thats why the point clouds were saved within 4 frames so that an evaluation will be done on the point cloud of 4 frames earlier. This way we prevent the algorithm to use a fuzzy point cloud in case the user head movement too fast making the point cloud inaccurate. The main concern of our training process was overfitting due to the small dataset. For improved results a larger dataset has to be created.

\section{CONCLUSION}
We proposed a flexible, intuitive method for Robot AR calibration using state of the art neural networks based only on 3D sensor data. We showed the feasibility of our method and could eliminate any external tools like markers. The method is applicable on any setup containing depth sensors. 
However, it struggles with real time object detection due to the intensive point cloud processing operations which is done within the AR device. An outsourcing of these processes inside a cloud for example would accelerate the process. Another aspect could be improved hardware or optimized algorithms for point cloud extraction.another downside is that we only trained the network with a small dataset. the accuracy could be improved with a larger dataset which has a variety of backgrounds or different robots. Nevertheless, our work show the feasibility of point cloud detection within AR device depth sensors and opens the door for further research. Furthermore, a 3D annotation tool has been proposed for work with raw point cloud data. 

\addtolength{\textheight}{-10cm}   % This command serves to balance the column lengths

%%%%%%%%%%%%%%%%%%%%%%%%%%%%%%%%%%%%%%%%%%%%%%%%%%%%%%%%%%%%%%%%%%%%%%%%%%%%%%%%

%%%%%%%%%%%%%%%%%%%%%%%%%%%%%%%%%%%%%%%%%%%%%%%%%%%%%%%%%%%%%%%%%%%%%%%%%%%%%%%%

%%%%%%%%%%%%%%%%%%%%%%%%%%%%%%%%%%%%%%%%%%%%%%%%%%%%%%%%%%%%%%%%%%%%%%%%%%%%%%%%

%%%%%%%%%%%%%%%%%%%%%%%%%%%%%%%%%%%%%%%%%%%%%%%%%%%%%%%%%%%%%%%%%%%%%%%%%%%%%%%%

\bibliographystyle{IEEEtran}

\bibliography{references}

\end{document}